\documentclass[twocolumn]{article}

\usepackage{arxiv_two_col}

\usepackage[utf8]{inputenc} % allow utf-8 input
\usepackage[T1]{fontenc}    % use 8-bit T1 fonts
\usepackage{hyperref}       % hyperlinks
\usepackage{booktabs}       % professional-quality tables
\usepackage{amsfonts}       % blackboard math symbols
\usepackage{nicefrac}       % compact symbols for 1/2, etc.
\usepackage{microtype}      % microtypography
\usepackage{lipsum}		% Can be removed after putting your text content
\usepackage{graphicx}
\usepackage{natbib}
\usepackage{doi}

\usepackage{wrapfig}
\usepackage{subcaption}
\usepackage{bbm}
\usepackage{amsmath}
\usepackage{tikz}
\usepackage{xurl}
\usepackage{pgfplots}
\pgfplotsset{compat=newest}
\usetikzlibrary{arrows.meta}

\graphicspath{ {graphics/} }

\title{Investigation into the Training Dynamics of Learned Optimizers}

\date{} 					% Or removing it

\author{
    Jan Sobotka\\
    Faculty of Information Technology\\
    Czech Technical University in Prague\\
    Thákurova 9, Prague, Czech Republic \\
    \texttt{sobotj11@fit.cvut.cz} \\
    \And
    Petr Šimánek\\
    Faculty of Information Technology\\
    Czech Technical University in Prague\\
    Thákurova 9, Prague, Czech Republic \\
    \texttt{simanpe2@fit.cvut.cz} \\
    \And
    Daniel Vašata\\
    Faculty of Information Technology\\
    Czech Technical University in Prague\\
    Thákurova 9, Prague, Czech Republic \\
    \texttt{daniel.vasata@fit.cvut.cz} \\
}

\renewcommand{\shorttitle}{Investigation into the Training Dynamics of Learned Optimizers}

\hypersetup{
    pdftitle={Investigation into the Training Dynamics of Learned Optimizers},
    pdfauthor={Jan Sobotka, Petr Šimánek, Daniel Vašata},
    pdfkeywords={Learning to Optimize, Meta-Learning, Optimization},
}

\begin{document}
\twocolumn[\maketitle]

\begin{abstract}
    Optimization is an integral part of modern deep learning. Recently, the concept of learned optimizers has emerged as a way to accelerate this optimization process by replacing traditional, hand-crafted algorithms with meta-learned functions. Despite the initial promising results of these methods, issues with stability and generalization still remain, limiting their practical use. Moreover, their inner workings and behavior under different conditions are not yet fully understood, making it difficult to come up with improvements. For this reason, our work examines their optimization trajectories from the perspective of network architecture symmetries and parameter update distributions. Furthermore, by contrasting the learned optimizers with their manually designed counterparts, we identify several key insights that demonstrate how each approach can benefit from the strengths of the other.
\end{abstract}

% keywords can be removed
% \keywords{Learning to Optimize \and Meta-Learning \and Optimization}

\section{Introduction}
\label{sec:introduction}
    In the last decade, the field of machine learning has undergone many trends, the most significant being the replacement of manual feature engineering with learned features. Relatively recently, a similar perspective has been applied to optimization, the driving force behind deep learning. Specifically, the subfield of meta-learning called \textit{learning to optimize} (L2O) has the ambitious goal of learning the optimization itself, more or less replacing the hand-engineered algorithm approach.

    There is a substantial body of literature devoted to learning to optimize with available gradient information. These works try to learn the "gradient descent-like" method to optimize the parameters (weights and biases) of neural networks. The approaches range from learning adaptive step size (learning rate) to learning the whole gradient descent algorithm with recurrent neural networks (\citet{10.5555/3157382.3157543}, \citet{lv2017learning}, \citet{metz2020tasks}, \citet{metz2021training}, \citet{Simanek_2022}).

    Compared to the relatively long history of hand-engineered optimizers, where extensive research into the theory, together with many empirical findings, informed numerous advances (momentum \citet{POLYAK19641}, AdaGrad \citet{JMLR:v12:duchi11a}, RMSProp \citet{tieleman2012lecture}, Adam \citet{kingma2017adam}), the field of learning to optimize is in its nascent stages. Many fundamental questions remain unanswered, and an extensive investigation into the training dynamics of learned optimizers is still lacking, which hinders well-informed development and further progress of the whole field. This is especially important since these methods have been shown to be brittle, difficult to scale, and relatively ineffective at generalizing across diverse problems.

    Additionally, it might be possible to improve manually engineered optimization algorithms by noticing which strategies L2O has found to be useful. In other words, research on hand-designed and learned optimizers can be mutually beneficial.

    In light of the aforementioned lack of understanding and our desire to improve traditional optimizers, we empirically study various properties of the optimization trajectories. Specifically, we 1) analyze the impact of symmetries introduced by network architectures; 2) examine the heavy-tailedness of noise in the predicted parameter updates; 3) investigate the update covariance, and lastly; 4) inspect the progression of the update size. 

    Our results show several major differences as well as similarities between the training dynamics under traditional and learned optimizers. Moreover, we notice close parallels to the recently proposed optimizer Lion \citep{chen2023symbolic} and shed more light on the strengths of these two approaches.

    In particular, our experiments demonstrate that similarly to Lion, learned optimizers break the geometric constraints on gradients that stem from architectural symmetries and that deviations from these constraints are significantly larger than those observed with previous optimizers like Adam or SGD. In the case of learned optimizers, we observe that a large deviation from these geometric constraints almost always accompanies the initial rapid decrease in loss during optimization. More importantly, regularizing against it severely damages performance, hinting at the importance of this freedom in L2O parameter updates.

    Furthermore, by studying the noise and covariance in the L2O parameter updates, we also demonstrate that, on the one hand, L2O updates exhibit less heavy-tailed stochastic noise, and, on the other hand, the variation in updates across different samples is larger.

    The paper is organized as follows. In Section \ref{sec:background}, learned optimizers and the Lion optimizer are introduced. Then Section \ref{sec:methods} describes the theoretical analysis of symmetries, gradient geometry, stochastic gradient noise, and update covariance. This, in turn, leads to a series of experiments presented in Section \ref{sec:experiments} followed by a discussion in Section \ref{sec:discussion}. Lastly, Section \ref{sec:related-work} connects our work to previous studies and shows promising parallels.

\section{Background}
\label{sec:background}
    Let us start by explaining the learning to optimize method, which is the main subject of our study, and then follow with the recently introduced Lion optimizer.

    \subsection{Learning to Optimize}
    \label{background-learning-to-optimize}
        The primary task is to minimize a given function $L(\theta)$ by optimizing its vector of parameters $\theta$. We focus primarily on first-order optimization methods, where at each step $t$ of the algorithm, the optimizer has access to the gradient $\nabla L(\theta_t)$ and suggests an update $g_t$ to get $\theta_{t+1}$ as
        \begin{equation}\label{eq:L2O-train}
            \theta_{t+1}=\theta_t + g_t.
        \end{equation}
        
        Our approach follows \citet{10.5555/3157382.3157543}. The core idea is to use a recurrent neural network $M$, parameterized by $\phi$, that acts as the optimizer. Specifically, at each time step $t$, this network takes its hidden state $h_t$ together with the gradient $\nabla L(\theta_t)$ and produces an update $g_t$ and a new hidden state $h_{t + 1}$,
        \begin{equation}\label{eq:L2O-update}
            [g_t, h_{t + 1}] = M\big(\nabla_{\theta} L(\theta_t), h_t, \phi\big).
        \end{equation}

        The sequence obtained by \eqref{eq:L2O-train} then aims to converge to a local minimum of $L$.
        Therefore, $M$ is called the optimizer, or meta-learner, and $L(\theta)$ is called the optimizee.

        The parameters $\phi$ of the optimizer $M$ are learned by stochastic gradient descent and updated every $u$-th training step where the hyper-parameter $u$ is called the unroll. The loss of the optimizer is the expectation of the weighted unrolled trajectory of the optimizee,
        \begin{equation}\label{eq:Optimizee-train}
            \mathcal{L}(\phi) = \mathbb{E}_L \sum_{\tau=1}^{u} w_{\tau} L(\theta_{\tau + ju - 1})
        \end{equation}
        where $\mathbb{E}_L$ is the expectation with respect to some distribution of optimizee functions $L$ and $w_{\tau}$ are weights that are typically set to 1. Additionally, $j$ denotes the number of previously unrolled trajectories; therefore, $(j+1)$-th unrolled trajectory corresponds to training steps $t = ju, ju + 1, ..., (j+1)u - 1$. In practice, there is often only one optimizee function $L$, so the expectation in \eqref{eq:Optimizee-train} is removed.

        In addition to updates of $\phi$ along a single optimizee training (an inner loop), there is also an outer loop where the entire optimizee training is restarted from some initial $\theta_{t=0}$ while the parameters $\phi$ continue to learn. In particular, the outer loop takes place only during the meta-training phase, where the main goal is to learn a good set of parameters $\phi$. Evaluation of the learned optimizer then happens in what is sometimes called meta-testing where $\phi$ is fixed and only the optimizee's parameters $\theta$ are updated.

\begin{figure}[t]
            \centering
            \includegraphics[width=1\linewidth]{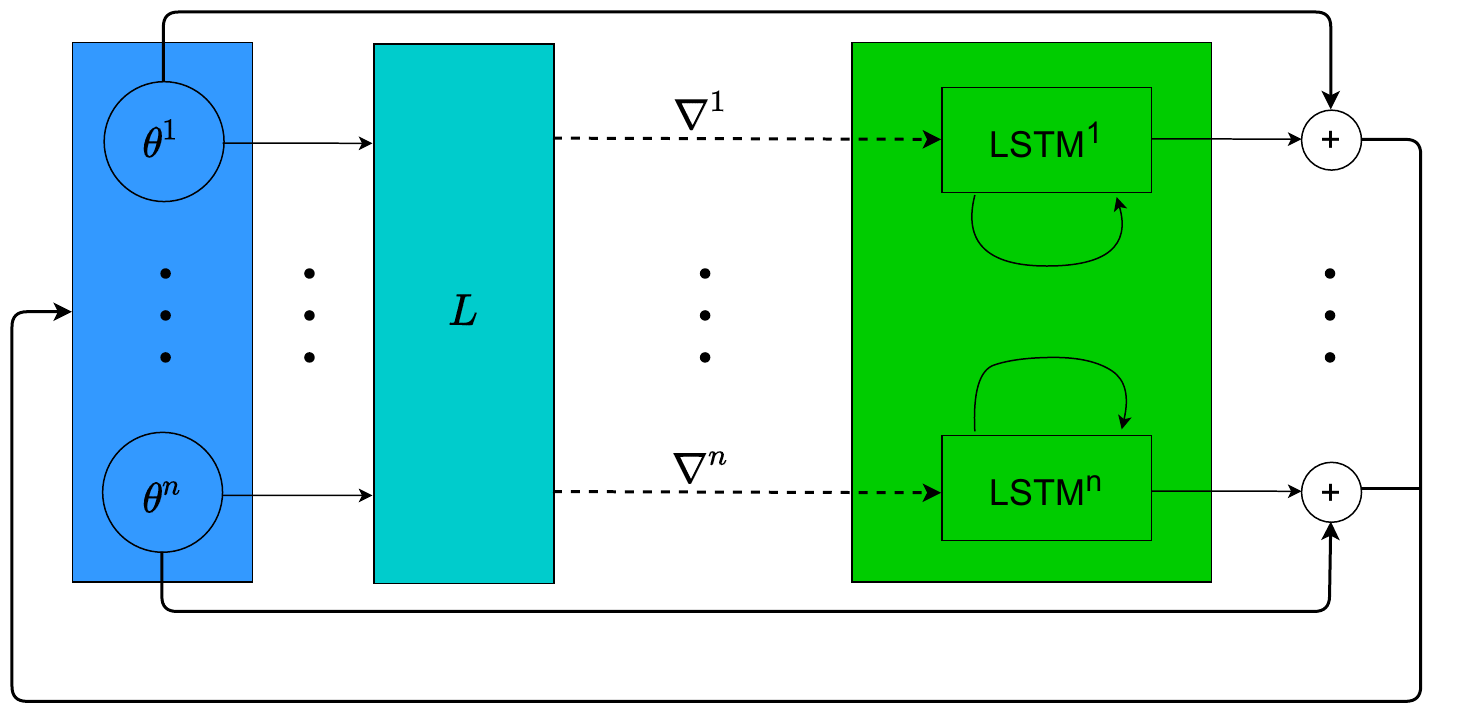}
        \caption{Single optimization step of L2O.}
        \label{fig:l2o-step}
        \end{figure}

        In practice, when there are several thousand or more parameters in $\theta$, applying a general recurrent neural network is almost impossible. The authors in \citet{10.5555/3157382.3157543} avoid this issue by implementing the update rule coordinate-wise using a two-layer LSTM network with shared parameters. This means that the optimizer $M$ is a small network with multiple instances operating on each parameter of the optimizee separately while sharing its parameters $\phi$ across all of them. A visualization of a single optimization step is presented in Figure \ref{fig:l2o-step}.

        We refer the reader to \citet{10.5555/3157382.3157543} for further algorithmic details and preprocessing.

    \subsection{Lion Optimizer}
        \label{background-lion}
        By applying program search techniques to the discovery of optimization algorithms, the authors of \citet{chen2023symbolic} came up with a simple yet highly effective adaptive algorithm called Lion (Evo\textbf{L}ved S\textbf{i}gn M\textbf{o}me\textbf{n}tum). The major distinction between Lion and other adaptive optimizers lies in its uniform update magnitude for each parameter, calculated through the sign operation. Remarkably, the authors demonstrated that this relatively simple optimization algorithm can outperform standard optimizers, such as Adam or Adafactor, across a wide range of benchmarks.

        A single step of the Lion optimizer starts with a standard weight decay step with a strength parameter $\lambda$. Then, Lion updates the parameters with the sign of the interpolation between the current gradient and momentum. The size of the step is determined by a learning rate $\eta$ and the interpolation is controlled by an exponential moving average (EMA) factor $\beta_1$. Finally, an update of the EMA is carried out using the EMA factor $\beta_2$.

        According to the authors, the sign operation in Lion introduces noise into the updates, which serves as a form of regularization. Empirical observations by the authors and previous studies \citet{foret2021sharpnessaware}, \citet{neelakantan2015adding} suggest that this noise can contribute to improved generalization. Additionally, their analysis indicated that Lion's updates generally exhibit a larger norm compared to those of Adam.

\section{Methods}
\label{sec:methods}
    \subsection{Symmetries and Gradient Geometry}
    \label{background-symmetries-in-network-architectures}
        In this section, we formalize the notion of differentiable symmetries, discuss the associated geometric properties of gradients in the context of neural networks, and present important examples.

        \subsubsection{Differentiable Symmetries}
        Symmetries of a function express its property of being invariant under a certain group of transformations. When these transformations are differentiable, we call them differentiable symmetries. In this paper, we consider symmetries of loss functions, and the subject of transformations in our case are the neural network's parameters.
        
        More formally, a function $f(\theta)$ where $\theta \in \mathbb{R}^n$ possesses a differentiable symmetry if it is invariant under the differentiable action $\psi: \mathbb R^n \times G \to \mathbb R^n$ of a group $G$ on the function's argument space. In other words, for all $\theta$ and all $\alpha \in G$, the output of the function does not change:
        \begin{equation}\label{eq:symmetry}
            f(\psi(\theta, \alpha)) = f(\theta).
        \end{equation}

        Note that our topic of interest is different from the work on equivariant neural networks, where the symmetry groups act on the input space, output space, and hidden feature spaces. Specifically, their focus is on the equivariance or invariance of $y = f_{\theta}(x)$ with respect to $x$ and $y$, but in the following sections, we are interested in symmetries acting on the parameter space instead.

        \subsubsection{Geometric Constraints on Gradients}
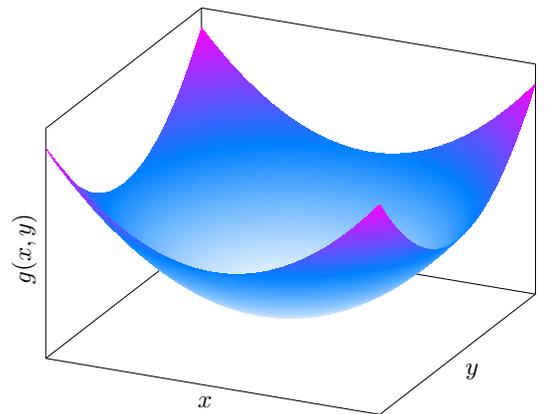
\begin{figure}[b]
            \centering
            \begin{tikzpicture}[scale=0.95]
                \begin{axis}[xlabel={$x$},ylabel={$y$},zlabel={$g(x,y)$},ticks=none,colormap/cool]
                    \addplot3[surf,shader=interp,domain=-1:1] {x^2 + y^2};
                \end{axis}
            \end{tikzpicture}
            \caption{Example function $g(x,y) = x^2 + y^2$.}
            \label{fig:quadratic}
        \end{figure}

        Let us start with an intuitive example to introduce the concept of gradient geometry. Consider the function $g(x,y) = x^2 + y^2$ depicted in Figure \ref{fig:quadratic} that possesses rotational symmetry
        \begin{equation}
            \psi(x,y,\alpha) =
            \begin{pmatrix}
                \cos{\alpha} & -\sin{\alpha}\\
                \sin{\alpha} & \cos{\alpha}
            \end{pmatrix}
            \begin{pmatrix}
                x\\
                y
            \end{pmatrix},
        \end{equation} 
        i.e. rotating around the origin by some angle $\alpha$ leaves the output of $g$ unchanged. 

        %%% Example function g
        
\begin{figure}[ht]
            \centering
            \begin{subfigure}[b]{0.33\linewidth}
                \centering
                \resizebox{\linewidth}{!}{
                    \begin{tikzpicture}
                        \begin{axis}[
                            xmin=-3, xmax=3,
                            ymin=-3, ymax=3,
                            zmin=0, zmax=1,
                            axis equal image,
                            view={0}{90},
                            xlabel={$x$},
                            ylabel={$y$},
                            label style={font=\LARGE},
                            ticks=none,
                        ]
                            \addplot3[
                                quiver={
                                    u={x/sqrt(x^2+y^2)},
                                    v={y/sqrt(x^2+y^2)},
                                    scale arrows=0.65,
                                    every arrow/.append style={%
                                        line width=.1+\pgfplotspointmetatransformed/1000,
                                        -{Latex[length=0pt 5,width=0pt 5]}
                                    },
                                },
                                -stealth,
                                samples=6,
                                domain=-3:3,
                                domain y=-3:3,
                                color=red,
                            ] {0};
                        \end{axis}
                    \end{tikzpicture}
                }
                % \caption{$\nabla g$}
                % \label{fig:example-grad}
            \end{subfigure}%
            \begin{subfigure}[b]{0.33\linewidth}
                \centering
                \resizebox{\linewidth}{!}{
                    \begin{tikzpicture}
                        \begin{axis}[
                            xmin=-3, xmax=3,
                            ymin=-3, ymax=3,
                            zmin=0, zmax=1,
                            axis equal image,
                            view={0}{90},
                            xlabel={$x$},
                            ylabel={$y$},
                            label style={font=\LARGE},
                            ticks=none,
                        ]
                            \addplot3[
                                quiver={
                                    u={-y/sqrt(x^2+y^2)},
                                    v={x/sqrt(x^2+y^2)},
                                    scale arrows=0.65,
                                    every arrow/.append style={%
                                        line width=.1+\pgfplotspointmetatransformed/1000,
                                        -{Latex[length=0pt 5,width=0pt 5]}
                                    },
                                },
                                -stealth,
                                samples=6,
                                domain=-3:3,
                                domain y=-3:3,
                                color=blue,
                            ] {0};
                        \end{axis}
                    \end{tikzpicture}
                }
                % \caption{$\partial_{\alpha} \psi \big|_{\alpha=0}$}
                % \label{fig:example-rotation}
            \end{subfigure}%
            \begin{subfigure}[b]{0.33\linewidth}
                \centering
                \resizebox{\linewidth}{!}{
                    \begin{tikzpicture}
                        \begin{axis}[
                            xmin=-3, xmax=3,
                            ymin=-3, ymax=3,
                            zmin=0, zmax=1,
                            axis equal image,
                            view={0}{90},
                            xlabel={$x$},
                            ylabel={$y$},
                            label style={font=\LARGE},
                            ticks=none,
                        ]
                            \addplot3[
                                quiver={
                                    u={x/sqrt(x^2+y^2)},
                                    v={y/sqrt(x^2+y^2)},
                                    scale arrows=0.65,
                                    every arrow/.append style={%
                                        line width=.1+\pgfplotspointmetatransformed/1000,
                                        -{Latex[length=0pt 5,width=0pt 5]}
                                    },
                                },
                                -stealth,
                                samples=6,
                                domain=-3:3,
                                domain y=-3:3,
                                color=red,
                            ] {0};

                            \addplot3[
                                quiver={
                                    u={-y/sqrt(x^2+y^2)},
                                    v={x/sqrt(x^2+y^2)},
                                    scale arrows=0.65,
                                    every arrow/.append style={%
                                        line width=.1+\pgfplotspointmetatransformed/1000,
                                        -{Latex[length=0pt 5,width=0pt 5]}
                                    },
                                },
                                -stealth,
                                samples=6,
                                domain=-3:3,
                                domain y=-3:3,
                                color=blue,
                            ] {0};
                        \end{axis}
                    \end{tikzpicture}
                }
                % \caption{$\big\langle \nabla g , \partial_{\alpha} \psi \big|_{\alpha=0} \big\rangle$}
                % \label{fig:example-dot-prod}
            \end{subfigure}
            \caption{Gradient geometry for the function $g(x,y) = x^2 + y^2$. Left: Directions of the gradient $\nabla g$. Middle: Directions of the vector field $\partial_{\alpha} \psi \big|_{\alpha=0}$ that generates the symmetry. Right: Orthogonality of the gradient and the symmetry-generating vector field.}
            \label{fig:geometric-constraints}
        \end{figure}
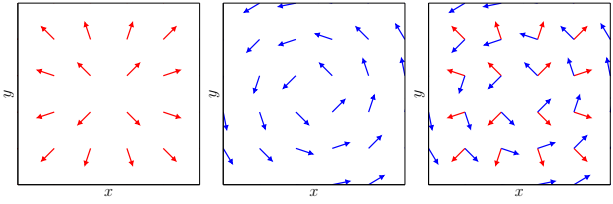

        Taking the partial derivative at the identity $\alpha = 0$, we get
        \begin{equation}
            \partial_{\alpha} \psi \big|_{\alpha=0} =
            \begin{pmatrix}
                -y\\
                x
            \end{pmatrix},
        \end{equation}
        where $\partial_{\alpha} \psi \big|_{\alpha=0}$ is the vector field that generates the symmetry. 
        Taking the gradient of $g$ at $(x,y)$ yields
        \begin{equation}
            \nabla g = 2
            \begin{pmatrix}
                x\\
                y
            \end{pmatrix}
        \end{equation}
        Hence, the gradient is perpendicular to the vector field:
        \begin{equation}
            \langle \nabla g, \partial_{\alpha} \psi \big|_{\alpha=0} \rangle = 0\quad\text{for all}\quad x,y.
        \end{equation}
        See Figure \ref{fig:geometric-constraints} for a depiction of this geometrical property.
        
        %%% illustration of a: func grad f; b: sym vector field; c: their overlap
        
        As will be now easily shown, such perpendicularity is not specific for the presented example, but it holds generally.

        Let us assume that $f$ is almost everywhere differentiable. By taking the derivative with respect to $\alpha$ at the identity $\alpha = I \in G$ of both sides of \eqref{eq:symmetry} and using the chain rule of differentiation, we get
        \begin{equation} \label{eq:grad-geometry}
            \big\langle \nabla f , \partial_{\alpha} \psi \big|_{\alpha=I} \big\rangle = 0,
        \end{equation}
        which implies that for all $\theta$, the gradient $\nabla f$ whenever exists is perpendicular to the vector field $\partial_{\alpha} \psi \big|_{\alpha=I}$ that generates the symmetry.
        
        Similar relationships hold for the Hessian matrices as can be found in \citet{kunin2021neural} together with formal proofs and a more detailed discussion of the above.

        Let us now introduce three families of symmetries commonly appearing in modern network architectures - translation, rescale, and scale. Since our analysis focuses on learning to optimize methods, we investigate how the predicted parameter updates adhere to the constraints on gradients. Moreover, we compare the deviations from these geometric constraints with the deviations of SGD, Adam, and Lion.

        Additionally, part of our analysis includes regularization of learned optimizers during meta-training against the aforementioned deviations. In such cases, we calculate the absolute deviations for each predicted update during meta-training using the optimizee's detached parameters and include these absolute deviations as part of the optimizer's loss. A description of the three symmetries and the method of calculating the deviations are provided below.

        \paragraph{Translation symmetry.}
        By translation symmetry we mean the invariance of a function to action $\psi(\theta, \alpha): \theta \mapsto \theta + \alpha \xi$, where $\xi \in \mathbb{R}^n$ is some constant vector and $\alpha \in \mathbb R$. We will consider only cases where $\xi = \mathbbm{1}_{X}$, which is the indicator vector for some subset $X$ of $\{1, ..., n\}$. An example is the vector $\mathbbm{1}_{\{1,3\}} = (1,0,1,0,\dotsc, 0)$. From \eqref{eq:grad-geometry}, we get
        \begin{equation}
            \big\langle \nabla f, \partial_{\alpha} \psi \big|_{\alpha=0} \big\rangle = \\
            \big\langle \nabla f, \mathbbm{1}_{X} \big\rangle = \\
            0.
        \end{equation}
        In other words, the gradient is orthogonal to the indicator vector. 
        
        Let us have a subset $X = \{i_1, i_2, \dotsc, i_k\}$ of $\{1, ..., n\}$ and denote by $\nabla_{\theta_X} f$ the gradient of $f$ with respect to a subset of its arguments given by $\theta_X = (\theta_{i_1}, \theta_{i_2},\dotsc, \theta_{i_k}) \in \mathbbm R^k$. Then, we may rewrite the previous orthogonality as
        \begin{equation}
            \label{eq:perpendicularity}
            \big\langle \nabla f, \mathbbm{1}_{X} \big\rangle =\\
            \big\langle \nabla_{\theta_X} f, \mathbbm{1} \big\rangle = \\
            0,
        \end{equation}
        where $\mathbbm{1} = (1,\dotsc, 1)$.

        In the context of deep learning, the translation symmetry is present in the softmax function $\sigma_i(z) = \frac{e^{z_i}}{\Sigma_j e^{z_j}}$ with $z = Wx + b$. Shifting any column $W_{:,i}$ or the bias vector $b$ by a real constant leaves the output of the softmax unchanged. Therefore, the softmax function is invariant under translation, and this symmetry induces from \eqref{eq:perpendicularity} the following gradient property:
        \begin{equation}
            \big\langle \nabla_{W_{:,i}} L, \mathbbm{1} \big\rangle = \\
            \big\langle \nabla_{b} L, \mathbbm{1} \big\rangle = \\
            0,
        \end{equation}
        where we denote by $L$ the loss function.

        The equation above provides us with a simple way to calculate absolute deviation from this symmetry invariance as follows:
        \begin{equation}
            \Big\lvert\big\langle g_{b}(\nabla L), \mathbbm{1} \big\rangle\Big\rvert +
            \sum_i
            \Big\lvert\big\langle g_{W_{:,i}}(\nabla L), \mathbbm{1} \big\rangle\Big\rvert,
        \end{equation} 
        where the function $g$ denotes the update rule of an optimizer with the corresponding components $g_{b}$, $g_{W_{:,i}}$, and the index $i$ iterates over all the columns of $W$.

        \paragraph{Rescale symmetry.}
        Rescale symmetry is in our case defined by the group $GL_{1}^{+}(\mathbb{R})$ and the action $\psi(\theta, \alpha): \theta \mapsto a_{X_1}(\alpha) \odot a_{X_2}(\alpha^{-1}) \odot \theta$, where $X_1$ and $X_2$ are two disjoint subsets of $\{1, ..., n\}$, $\alpha \in (0,+\infty)$, $a_{X_1}(\alpha) = \mathbbm{1} + (\alpha - 1)\mathbbm{1}_{X_1}$, and $\odot$ is the element-wise multiplication. An example of the notation is $a_{\{1,3\}}(\alpha) = (\alpha,1,\alpha,1,\dotsc, 1)$. 
        Function $f$ possesses rescale symmetry if $f(\theta) = f(\psi(\theta, \alpha))$ for all $\alpha \in GL_{1}^{+}(\mathbb{R})$. Taking similar steps as before, this symmetry implies gradient orthogonality as
        \begin{equation}
            \big\langle \nabla f, \partial_{\alpha} \psi \big|_{\alpha=1} \big\rangle = \\
            \big\langle \nabla f, \theta \odot \mathbbm{1}_{X_1} - \theta \odot \mathbbm{1}_{X_2} \big\rangle = \\
            0.
        \end{equation}
        Full formal proofs can be found in \citet{kunin2021neural}.

        Interestingly, the rescale symmetry is present at every hidden neuron of networks with continuous, homogeneous activation functions such as ReLU and Leaky ReLU. A simple illustration of this comes from considering a single neuron's computational path $w_2 \sigma(w_1^T x + b)$ and noticing that scaling $w_1$ and $b$ by a real constant and $w_2$ by its inverse has no effect since the constants can be passed through the activation function, canceling their effect. Thus, the gradient constraints in this case are
        \begin{equation}
            \big\langle \nabla_{w_1} L, w_1 \big\rangle + \\
            \big\langle \nabla_{b} L, b \big\rangle - \\
            \big\langle \nabla_{w_2} L, w_2 \big\rangle = \\
            0.
        \end{equation}

        Absolute deviation of the optimizer's updates given by $g$ from the rescale symmetry constraints can be therefore defined as
        \begin{equation}
            \Big\lvert\big\langle g_{w_1}(\nabla L), w_1 \big\rangle +
            \big\langle g_{b}(\nabla L), b \big\rangle -
            \big\langle g_{w_2}(\nabla L), w_2 \big\rangle\Big\rvert.
        \end{equation}

        \paragraph{Scale symmetry.}
        Scale symmetry can be defined by the group $GL_{1}^{+}(\mathbb{R})$ and the action $\psi(\theta, \alpha): \theta \mapsto a_X(\alpha) \odot \theta$, where the notation is the same as for the rescale symmetry. Given this definition, a function possesses scale symmetry if $f(\theta) = f(\psi(\theta, \alpha)) = f(\alpha_X \odot \theta)$ for all $\alpha \in GL_{1}^{+}(\mathbb{R})$. The gradient orthogonality arising from this symmetry is
        \begin{equation}
            \big\langle \nabla f, \partial_{\alpha} \psi \big|_{\alpha=1} \big\rangle = \\
            \big\langle \nabla f, \theta \odot \mathbbm{1}_{X} \big\rangle = \\
            \big\langle \nabla_{\theta_{X}} f, \theta_{X} \big\rangle = \\
            0
        \end{equation}
        i.e. the gradient of the function is everywhere perpendicular to the parameter vector itself.

        Batch normalization, as used in deep learning, has this scale invariance during training. To see this, consider the incoming weights $w \in \mathbb{R}^d$ and bias $b \in \mathbb{R}$ of a neuron with batch normalization $\frac{z - \bar z}{\sqrt{\text{var}\left(z\right)}}$, where $z = w^T x + b$, $\bar z$ is the sample average, and $\text{var}\left(z\right)$ is some standard estimation of variance. Scaling $w$ and $b$ by a non-zero real constant has no effect on the output since it factors from $z$, $\bar z$, and $\sqrt{\text{var}(z)}$ canceling the effect. Thus, $w$ and $b$ observe scale symmetry in the loss, and their gradients satisfy 
        \begin{equation}
            \langle \nabla_w L, w \rangle + \langle \nabla_b L, b \rangle = 0.
        \end{equation}
        More details can be found in \citet{ioffe2015batch} and \citet{kunin2021neural}.

        Therefore, we can calculate the absolute deviation of the updates given by $g$ from the scale symmetry constraints as
        \begin{equation}
            \Big\lvert\langle g_{w}(\nabla L),\ w \rangle + \langle g_{b}(\nabla L),\ b \rangle\Big\rvert.
        \end{equation}
    \subsection{Stochastic Gradient Noise and \texorpdfstring{$\alpha$}{α}-Stable Distributions}
    \label{background-grad-noise-alpha-distributions}
        \subsubsection{Stochastic Gradient Noise}
        Stochastic gradient noise refers to the fluctuations in the gradient of the loss function. 
        The gradient noise vector is defined by
        \begin{equation}
            U_k = \nabla L_k - \nabla L,
        \end{equation}
        where $\nabla L_k$ is the $k$-th mini-batch gradient and $\nabla L$ is the full-batch gradient of the loss function.
        
        Although many theoretical works have assumed $U_k$ to be Gaussian with independent components, the authors of \citet{pmlr-v97-simsekli19a} have shown that it is not entirely the case. Their theory suggested that the gradient noise converges to a heavy-tailed $\alpha$-stable random variable and proposed a novel analysis of SGD, which their experiments showed to be more suitable. Specifically, in a series of experiments with common deep learning architectures, the gradient noise exhibited a highly non-Gaussian and heavy-tailed distribution.

        More importantly, they proved that gradient noise with heavier tails increases the probability of ending up in a wider basin of the loss landscape, an indication of better generalization (\cite{10.1162/neco.1997.9.1.1}).
        
        \subsubsection{Symmetric $\alpha$-Stable Distributions}
        One can view the symmetric $\alpha$-stable ($S \alpha S$) distribution as a heavy-tailed generalization of a centered Gaussian distribution. The $S \alpha S$ distributions are defined through their characteristic function
        \begin{equation}
            X \sim S \alpha S(\sigma) \Longleftrightarrow \mathbb{E} \big[ e^{i \omega X} \big] = e^{-|\sigma \omega|^{\alpha}}.
        \end{equation}

        In general, the probability density function does not have a closed-form formula, but it is known that the density decays with a power law tail like $|x|^{-\alpha - 1}$ where $\alpha \in (0, 2]$ is called the tail-index: as $\alpha$ gets smaller, the distribution has a heavier tail. 

        Under the assumption of the same tail-index for all components of the gradient noise update and of the parameter update noise defined by
        \begin{equation}
            U_k = g_k - g,
        \end{equation}
        where $g_k$ is $k$-th mini-batch parameter update and $g$ is the the full-batch update, we estimate the tail-index $\alpha$ of distribution of the gradient noise components and parameter update noise components using the estimator introduced in \citet{RePEc:spr:metrik:v:78:y:2015:i:5:p:549-561} and specifically its implementation from \citet{pmlr-v97-simsekli19a}.

        %The additional parameter $\sigma \in \mathbb{R}_+$, also known as the scale parameter, controls the spread of the random variable $X$ around 0. In fact, setting $\alpha=2$, the $S \alpha S$ turns into the Gaussian distribution $\mathcal{N}(0,2\sigma^2)$.

    \subsection{Update Covariance}
    \label{methods-covariance}
        To further investigate the noise in the mini-batch parameter updates for different optimizers, we study the magnitude of their variation across different samples. We do so by considering the update covariance matrix
        \begin{equation}
            K = \frac{1}{N} \sum_{i=0}^N (g_i - g) (g_i - g)^T,
        \end{equation} where $g_i$ is the parameter update in the sample $x_i$, $g$ is the full-batch update, and $N$ is the number of training samples.

        Computing and storing $K$ in memory is expensive due to the quadratic cost in the number of parameters. However, since we want to analyze the magnitude of the deviations of the parameter updates from the full-batch update, it is sufficient to determine only the largest eigenvalue of $K$.
        
        To address this, we use the mini-batch updates as follows. We begin by sampling $L$ mini-batch update of size $M$ and compute the corresponding $L\times L$ Gram matrix $K^M$ with entries
        \begin{equation}
            K_{i,j}^M = \frac{1}{L} \langle g_i - g, g_j - g \rangle,
        \end{equation}
        where $g_i$ is $i$-th mini-batch update and $g$ is the estimation of the full-batch update based on the $L$ mini-batches (in our analysis, we use $L=93$). Subsequently, we find the maximum eigenvalue of $K^M$ using the power iteration method. The empirical findings of the authors of \citet{Jastrzebski2020The} demonstrate that the largest eigenvalue of $K^M$ approximates the largest eigenvalue of $K$ quite well.

\section{Experiments}
\label{sec:experiments}
    All our meta-training is performed on feed-forward neural network optimizees with 1 hidden layer of 20 neurons with either sigmoid, Leaky ReLU, or batch normalization followed by ReLU. We put the softmax activation function at the output and train the optimizees on the MNIST classification task with the cross-entropy loss function and batch size of 128. We set unroll to 20 iterations, learning rate to 0.001, and perform meta-training for 50 epochs, each consisting of 20 separate optimization runs with the maximum iteration number of 200.

    During meta-testing, we evaluate L2O on the optimizee architectures from meta-training as well as on larger networks described in each experiment separately.

    For SGD, Adam, and Lion, by performing a hyperparameter search, we chose a learning rate of 0.1 and momentum of 0.9 for SGD, a learning rate of 0.05 for Adam, and learning rates 0.001, 0.005, 0.01 for sigmoid, Leaky ReLU, and ReLU with batch normalization, respectively, for Lion.

    \subsection{Breaking the Geometric Constraints}
    \label{experiments-breaking-the-geometric-constraints}
        To assess the importance of learned optimizers being free from the geometric constraints that might bind classical optimizers, we follow a two-step procedure. First, we measure the deviations of L2O from the aforementioned geometric constraints for the translation, rescale, and scale symmetries. Second, we regularize the learned optimizer during meta-training against breaking these constraints and observe its effects on the performance during meta-testing. The goal is to better understand the difference between the optimization trajectories of L2O and classical optimizers through the lens of symmetries in the optimizee architecture.

        \paragraph{Deviations from the constraints.}
        We meta-train and subsequently meta-test L2O on the aforementioned optimizee models and track the progression of the deviations of parameter updates from the geometric constraints. Additionally, we compare the deviations of hand-engineered optimizers such as SGD, Adam, and Lion.

\begin{figure}[t]
            \centering
            \includegraphics[width=\linewidth]{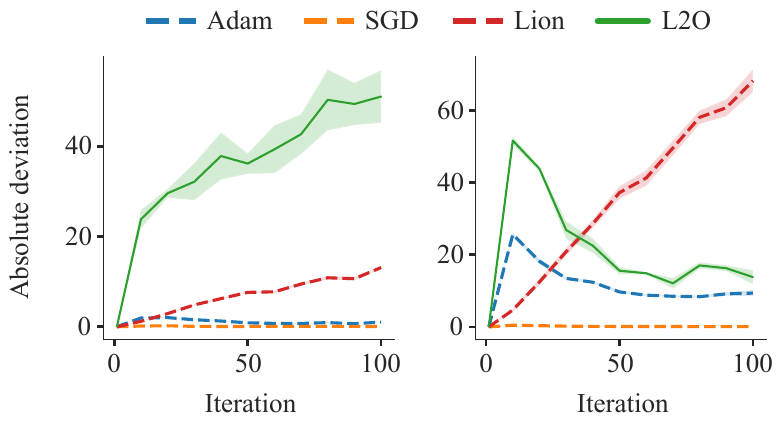}
            \caption{Deviations from the geometric constraints. Left: Rescale symmetry breaking on the Leaky ReLU optimizee. Right: Scale symmetry breaking on the optimizee with batch normalization and ReLU.}
            \label{fig:sym_breaking}
        \end{figure}

        As can be seen in Figure \ref{fig:sym_breaking}, where L2O was meta-trained on optimizees with Leaky ReLU and ReLU with batch normalization, the deviations from the geometric constraints from the rescale and scale symmetries are mostly larger than those of SGD and Adam. But most strikingly, the increase in this symmetry breaking is largest for L2O at the beginning of optimization, whereas for Lion, it increases more gradually and achieves higher values later.
        
        %%% Symmetry breaking - deviations
        
\begin{figure}[b]
            \centering
            \includegraphics[width=\linewidth]{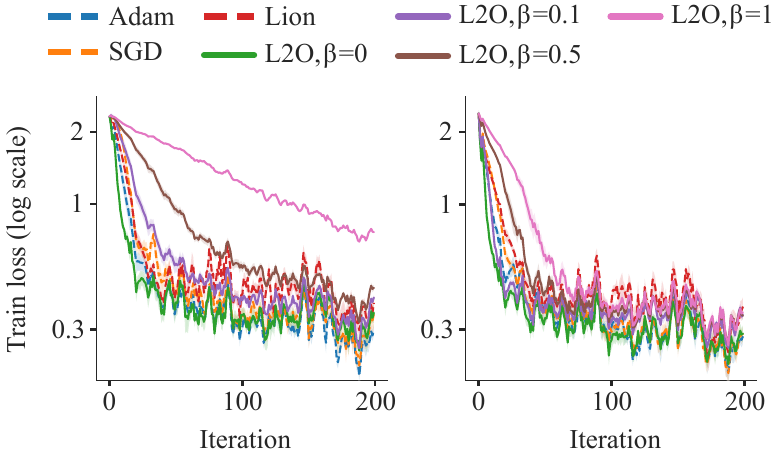}
            \caption{Performance after the symmetry breaking regularization. Left: Rescale symmetry breaking regularized on the Leaky ReLU optimizee. Right: Scale symmetry breaking regularized on the optimizee with ReLU and batch normalization.}
            \label{fig:regularization_effects_on_performance}
        \end{figure}

        \paragraph{Symmetry breaking regularization.}
        To get a deeper insight into how L2O leverages the freedom of parameter updates, we meta-train with an additional regularization loss that penalizes the absolute size of the L2O's update deviations from the geometric constraints on gradients. The performance for various regularization strengths $\beta$ is shown in Figure \ref{fig:regularization_effects_on_performance}.

        Interestingly, as regularization increases, the L2O's optimization speed significantly drops. This observation reiterates the interesting L2O characteristic that its optimization trajectories do not escape the symmetry-induced equivalent parameter combinations as gradient would, but rather take a route more aligned with the level sets of the loss landscape. Such a search for a better starting point for the next step might be related to the symmetry teleportation presented in \citet{zhao2023symmetry}, where the authors performed loss-invariant symmetry transformations of the parameters to find combinations with a steeper gradient. Given that L2O is meta-trained and its predicted parameter updates are history-dependent, it is an interesting topic for future study to analyze this connection further.
        
\begin{figure*}
        \centering
            \begin{subfigure}[b]{0.33\linewidth}
                \centering
                \includegraphics[width=\linewidth]{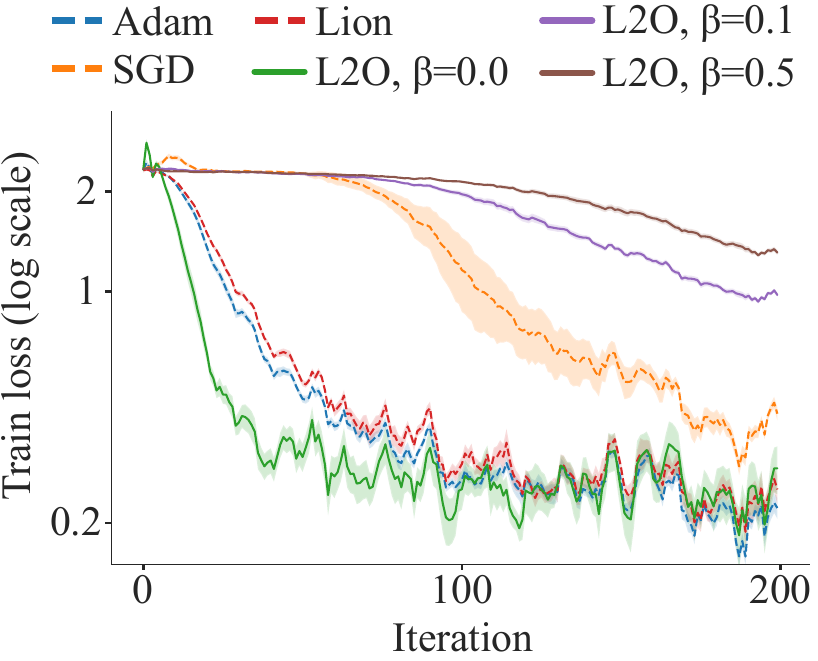}
            \end{subfigure}%
            \begin{subfigure}[b]{0.33\linewidth}
                \centering
                \includegraphics[width=\linewidth]{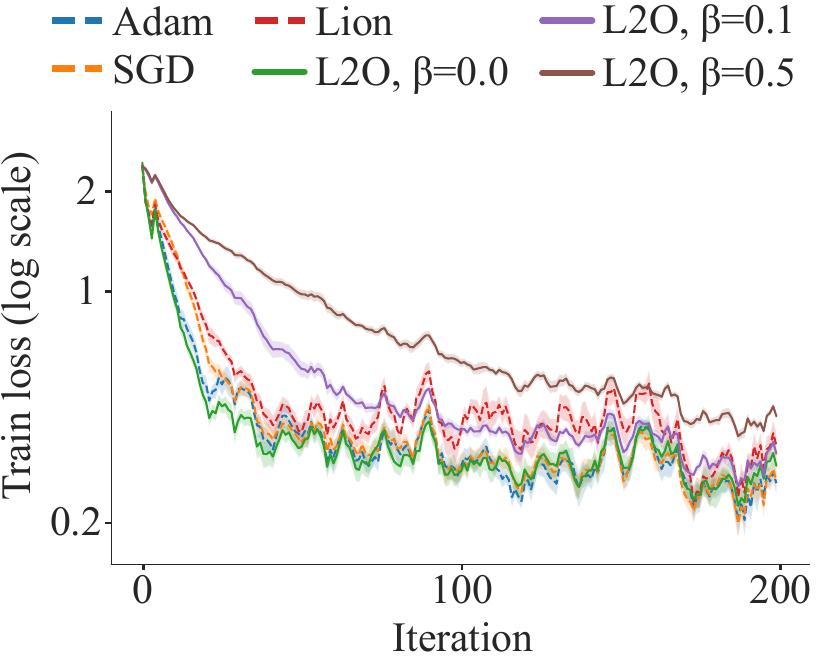}
            \end{subfigure}%
            \begin{subfigure}[b]{0.33\linewidth}
                \centering
                \includegraphics[width=\linewidth]{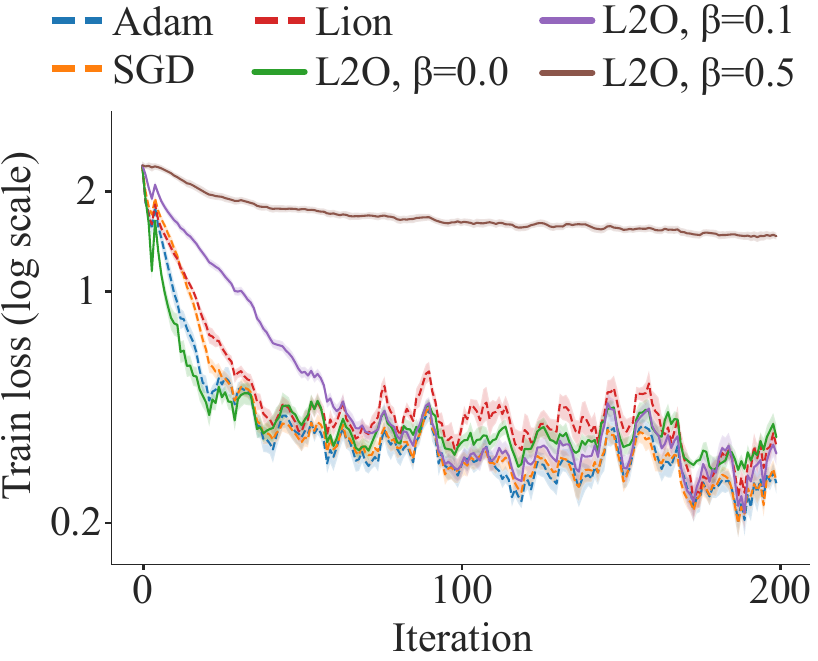}
            \end{subfigure}
        \caption{Effects of regularization strength $\beta$ on the performance of L2O on novel optimizee architectures. Left: L2O meta-trained on the Leaky ReLU optimizee with regularization against the rescale symmetry breaking and then meta-tested on 5x wider and 2x deeper optimizee network with sigmoid. Middle: L2O meta-trained on the Leaky ReLU optimizee with regularization against the rescale symmetry breaking and then meta-tested on ReLU with batch normalization. Right: L2O meta-trained on the sigmoid optimizee with regularization against the translation symmetry breaking and then meta-tested on ReLU with batch normalization.}
        \label{fig:regularization-effects-on-performance-novel-optees}
        \end{figure*}

        The same observation for the effect of regularization can be made for most of the optimizee architectures on which L2O was not meta-trained (Figure \ref{fig:regularization-effects-on-performance-novel-optees}). Results for L2O meta-trained on the optimizee with sigmoid activation function can be found in the Appendix.
        
        %%% Effects of regularization on performance - novel optimizees
        
        \paragraph{Symmetry breaking and performance of traditional optimizers.}
        We can also relate the symmetry breaking of classical hand-designed optimizers to their performance. Specifically, by interpolating between the optimization steps of SGD and Lion, we analyze how the change in symmetry breaking of this combined optimizer coalesces with the change in its performance.
        
\begin{figure}[!b]
            \centering
            \includegraphics[width=\linewidth]{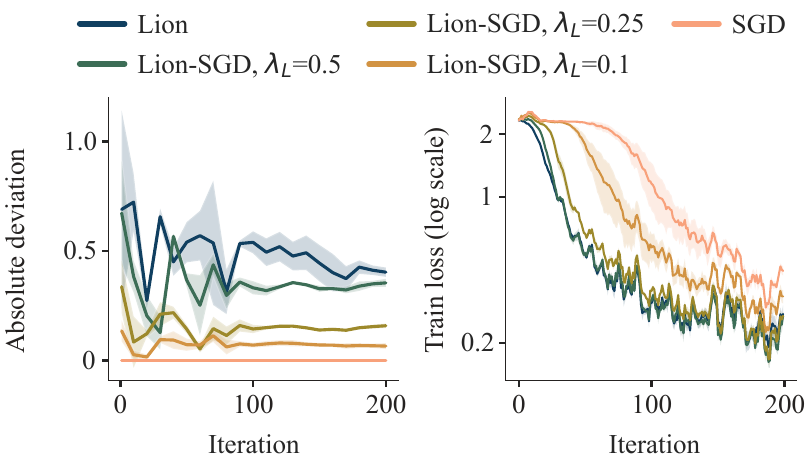}
            \caption{Deviations from the geometric constraints and the performance of the Lion-SGD optimizer. Left: Translation symmetry breaking on the sigmoid optimizee from Figure \ref{fig:regularization-effects-on-performance-novel-optees} on the left. Right: Training loss on the same optimizee.}
            \label{fig:sym_breaking_lion_sgd}
        \end{figure}

        Figure \ref{fig:sym_breaking_lion_sgd} shows how the Lion-SGD interpolation maps onto the deviations from the symmetry constraints (left) and the performance (right). $\lambda_L$ refers to the interpolation coefficient in \begin{equation}
            g_{\text{Lion-SGD}} = \lambda_L \cdot g_{\text{Lion}} + (1 - \lambda_L) \cdot g_{\text{SGD}},
        \end{equation}
        where $g$'s denote the parameter updates of the corresponding optimizers. The optimizee, in this case, is a feed-forward neural network with 2 hidden layers of 100 neurons with sigmoid activation function (the same as in Figure \ref{fig:regularization-effects-on-performance-novel-optees} on the left).

        Similarly to the earlier observation that the regularization of L2O symmetry breaking hinders its performance, we see that the increasing symmetry breaking of Lion-SGD correlates with an increase in performance. This indicates that breaking the strict geometric constraints is beneficial not only for L2O but also for more traditional, manually designed optimization algorithms.
        
    \subsection{Heavy-Tailed Distribution of Gradient and Parameter Update Noise}
    \label{experiments-heavy-tail-grads-updates}
        In this section, we investigate the progression of the estimated tail-index $\alpha$ of the gradient noise components and the parameter update noise components for different optimizers. With lower values of $\alpha$ indicating a more heavy-tailed distribution of noise of gradients and updates, we are particularly interested in comparing how the suggested parameter updates from SGD, Adam, and L2O evolve over a training run.

        The results for L2O meta-trained on the Leaky ReLU optimizee are shown in Figure \ref{fig:grad_update_alpha_cov} (left).

        %%% Symmetry breaking & performance - Lion-SGD
        
        First, we see that the estimates of $\alpha$ for L2O parameter updates are much higher than for the gradients, implying that the distribution of the noise in its updates is less heavy-tailed than that of the gradient. This shows that L2O effectively attenuates the heavy-tail portion of deviations in the gradient estimates on its input, taking a less jittery optimization trajectory. Moreover, one can see that except for a few drops of $\alpha$ at the beginning of training, the noise in the updates from L2O is generally less heavy-tailed than the updates from Adam or SGD. Given the previous studies (\citet{pmlr-v97-simsekli19a}) on the benefits of stochastic gradient noise for generalization, future developments of the L2O methods might want to take this observation of less heavy-tailed update noise into account when troubleshooting optimizee generalization gaps. Since our experiments are performed on a rather simple task of MNIST classification, we do not encounter such problems. Results for the other settings can be found in the Appendix.
        
        Second, since L2O was meta-trained only on training runs of 200 iterations, we observed that the loss in the meta-testing on sigmoid and ReLU with batch normalization optimizees starts to increase after around the 500th iteration. This inability to generalize to longer training sequences accurately correlates with decreasing $\alpha$ estimates for L2O's parameter updates and gradients on these two optimizee architectures as can be seen in the Appendix. With the Leaky ReLU optimizee network in Figure \ref{fig:grad_update_alpha_cov}, the training loss does not increase as for the other two optimizee architectures and so the $\alpha$ estimates are kept at around the same value even after the 500th iteration.
        
    \subsection{Update Covariance}
    \label{experiments-grads-updates-covariance}
        As described in Methods (Section \ref{methods-covariance}), we want to investigate the noise (variation) in the mini-batch parameter updates for different optimizers. We follow the same setup as in the previous experiments. The results for update covariance of all considered optimizers on the Leaky ReLU optimizee are shown in Figure \ref{fig:grad_update_alpha_cov} (right). Results for the other optimizees can be found in the Appendix.

        Interestingly, once again, we observe a similar relative ordering among the considered optimizers. In this case, the variation in updates across different samples appears to be the lowest for SGD, followed by Adam, Lion, and, finally, the largest variation is observed for L2O.

        %%% Heavy-tailed update noise + update covariance
        \begin{figure}[h]
            \centering
            \includegraphics[width=\linewidth]{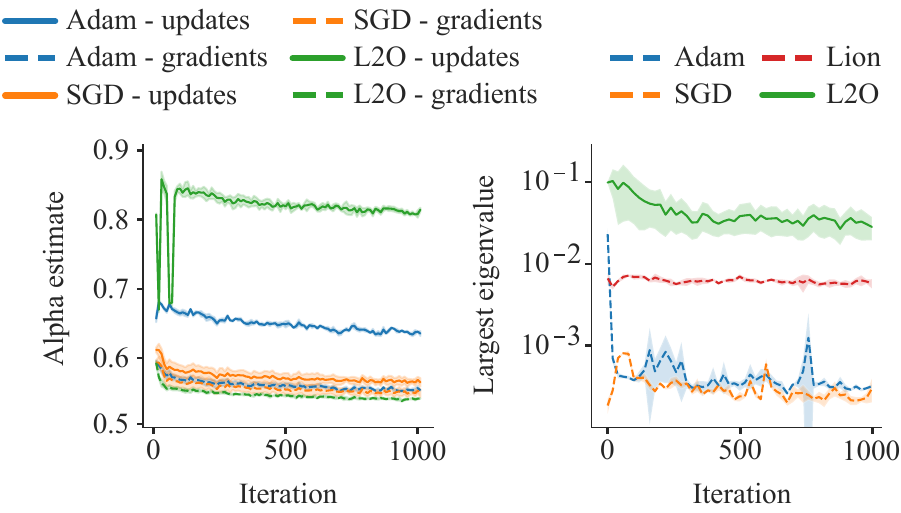}
            \caption{Heavy-tailedness and update covariance. Left: $\alpha$ estimates for the gradient and update noise on the Leaky ReLU optimizee. Right: Largest eigenvalue of the update covariance on the Leaky ReLU optimizee.}
            \label{fig:grad_update_alpha_cov}
        \end{figure}

    \subsection{Update Histograms}
    \label{experiments-grads-updates-histograms}
\begin{figure}[t]
            \centering
            \includegraphics[width=0.92\linewidth]{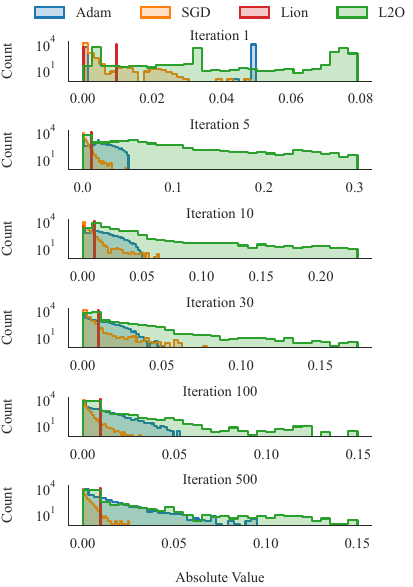}
        \caption{Histograms of the absolute values of parameter updates of different optimizers on the sigmoid optimizee.}
        \label{fig:update-hist-sigmoid}
        \end{figure}

        To compare the strategies of different optimization algorithms further, we examine the absolute values of their updates, as shown in Figure \ref{fig:update-hist-sigmoid}.

        One can notice that the L2O starts with the largest updates and then slowly approaches the update distribution of Adam. These large initial updates closely parallel the rapid symmetry breaking of the learned optimizer at the beginning of the optimization run. Such behavior holds also for the other optimizees we considered (in the Appendix).
        
        %%% Update histograms

\section{Discussion}
\label{sec:discussion}
    We found that one of the most pronounced features of learned optimizers is their rapid symmetry breaking at the beginning of the optimization run. Remarkably, the good performance of L2O in the initial phase of training correlates with this behavior very well, as is also demonstrated by the symmetry-breaking regularization, which heavily hindered the optimizer.

    Another aspect is the less heavy-tailed distribution of L2O updates despite the gradients exhibiting very heavy-tailed behavior. Together with the high variation of updates across different samples, as shown by large maximum eigenvalues of update covariance, this points to one interesting observation: L2O appears to act as a stabilizing force in the optimization process. While the inherent stochasticity and heavy-tailed nature of gradients might lead to erratic updates and slow convergence, the noise clipping of L2O seems to mitigate these issues.

\section{Related Work}
\label{sec:related-work}
    Since this paper focuses mainly on the investigation of the inner workings and behavior of L2O, we restrict the scope of this section to other similar works that attempt to analyze L2O.
    
    In \citet{harrison2022closer}, the authors used dynamical systems tools in a noisy quadratic model to study the stability of black-box optimizers. Their findings led them to propose a model they called \textit{stabilized through ample regularization} (STAR) learned optimizer. The authors argued that designing similar inductive biases informed by well-grounded theory and previous analysis is a potentially exciting and promising future direction. Closely related is the study of instability in the meta-training phase of L2O presented in \citet{metz2019understanding}.

    Similarly to our comparison of L2O with hand-designed optimizers, \citet{NEURIPS2021_a57ecd54} found that L2O has a relatively interpretable behavior. Specifically, they identified four standard algorithmic techniques: momentum, gradient clipping, learning rate schedules, and learning rate adaptation.

    In comparison to the above work which primarily studied the L2O's hidden state dynamics, we take a more high-level perspective and look at the optimization trajectories as a whole.

    Important to our paper is \citet{kunin2021neural} where the authors present a unified theoretical framework to understand the dynamics of neural network parameters during training with hand-designed optimizers. They based their study on intrinsic symmetries embedded in a network’s architecture that are present for any dataset and that impose stringent geometric constraints on gradients and Hessians.

    The connection between symmetries and optimization was also explored in the work of \citet{bamler2018improving}, where they showed that representation learning models possess an approximate continuous symmetry that leads to a slow convergence of gradient descent. This observation prompted them to introduce an optimization algorithm called Goldstone Gradient Descent, which aims to overcome this issue through continuous symmetry breaking. The core of their algorithm involves alternating between standard gradient descent and a specialized traversal in the subspace of symmetry transformations.

\section{Conclusion}
\label{sec:conclusion}
    Our investigation of the parameter update dynamics of the learned optimizers revealed several intriguing ingredients and strategies used by these meta-learning methods. Furthermore, our comparison with hand-engineered optimization algorithms has not only shown clear differences from traditional optimizers like SGD and Adam, but also illuminated similarities with the recently introduced method known as Lion. We believe that these findings pave the way for promising future research directions, where the insights gleaned from these side-by-side examinations can inform the design of more robust, scalable, and faster optimizers.

\section*{Acknowledgements}
    This work was supported by the Student Summer Research Program 2023 of FIT CTU in Prague.

\bibliographystyle{unsrtnat}
\bibliography{references}  %%% Uncomment this line and comment out the ``thebibliography'' section below to use the external .bib file (using bibtex) .

\clearpage
% \newpage
\appendix
\section{Appendix}
    \subsection{Additional Results}
\begin{figure}[h]
            \centering
            \includegraphics[width=\linewidth]{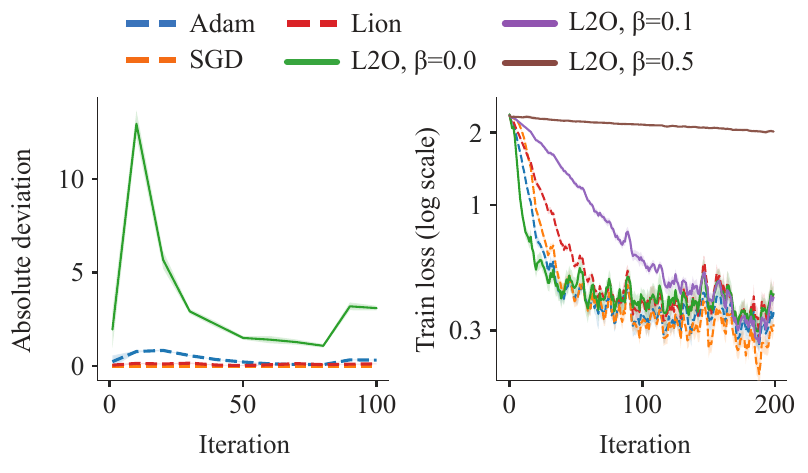}
            \caption{Symmetry breaking. Left: Translation symmetry breaking on the optimizee with sigmoid. Right: Performance after the translation symmetry breaking regularization on the optimizee with sigmoid.}
            \label{fig:sym_breaking_appendix}
        \end{figure}

        \paragraph{Symmetry breaking.}
        Figure \ref{fig:sym_breaking_appendix} shows the parameter update deviations from the constraints on gradients induced by translation symmetry, along with the associated performance after applying the symmetry breaking regularization.
        
        We see similar trends as for the Leaky ReLU and batch normalization optimizees. Specifically, L2O breaks the constraints by a large margin early in the training run.

        %%% Symmetry breaking - deviations and regularization performance
        
\begin{figure}[!b]
            \centering
            \includegraphics[width=\linewidth]{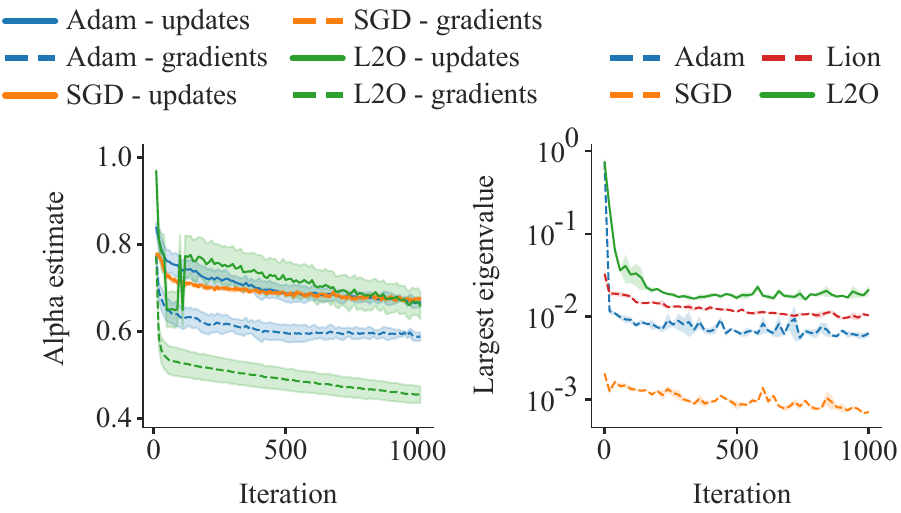}
            \caption{Heavy-tailedness and update covariance. Left: Gradient and update noise on the sigmoid optimizee. Right: Update covariance on the sigmoid optimizee.}
            \label{fig:grad_update_alpha_cov_appendix_sigmoid}
        \end{figure}

\begin{figure}
            \centering
            \includegraphics[width=\linewidth]{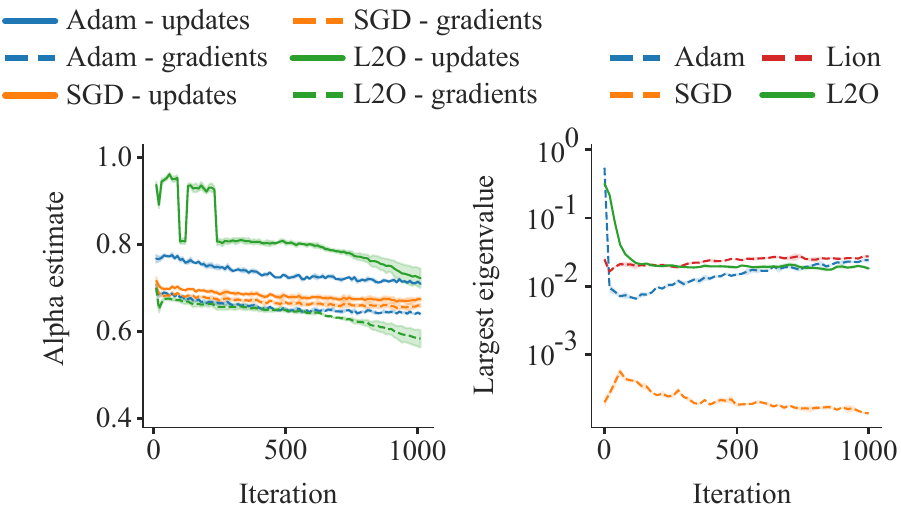}
            \caption{Heavy-tailedness and update covariance. Left: Gradient and update noise on the optimizee with batch normalization and ReLU. Right: Update covariance on the optimizee with batch normalization and ReLU.}
            \label{fig:grad_update_alpha_cov_relubn}
        \end{figure}

        \paragraph{Heavy-tailedness and update covariance.}
        The $\alpha$ estimates for the gradient and update noise, as well as the progression of the largest eigenvalue of the update covariance, are shown in Figures \ref{fig:grad_update_alpha_cov_appendix_sigmoid} and \ref{fig:grad_update_alpha_cov_relubn}.

        %%% Heavy-tailed update noise + update covariance - sigmoid
        
        Similarly to the results for the Leaky ReLU optimizee in the main part of this paper, we can again observe that L2O dampens the heavy-tailed gradient noise into high-variation parameter updates. Also, as mentioned earlier, we observed that the loss during L2O meta-testing on sigmoid and batch normalization optimizees starts to increase after around the 500th iteration and this accurately correlates with the decreasing $\alpha$ estimates of L2O gradient and update noise we see here.
        
        Additionally, an interesting observation in itself, we found that standardizing the image data has a relatively large impact on how heavy-tailed the gradient and update noise distributions are. Specifically, when we replaced the data normalization to the range $[0,1]$ with standardization, the alpha estimates shifted around 0.4 lower. Since we tried to follow the setup from \citet{10.5555/3157382.3157543}, all our results were obtained using data normalization.

        %%% Heavy-tailed update noise + update covariance - relu bn
        
\begin{figure}
            \centering
            \includegraphics[width=0.85\linewidth]{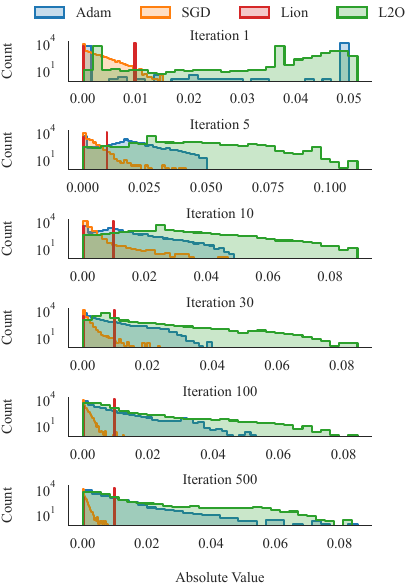}
        \caption{Histograms of the absolute values of parameter updates for L2O meta-trained and meta-tested on the optimizee with batch normalization and ReLU.}
        \label{fig:update_hist_relubn}
        \end{figure}

\begin{figure}
            \centering
            \includegraphics[width=0.85\linewidth]{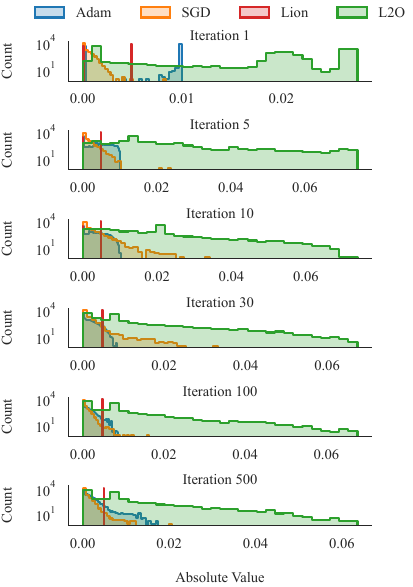}
        \caption{Histograms of the absolute values of parameter updates for L2O meta-trained and meta-tested on the Leaky ReLU optimizee.}
        \label{fig:update_hist_lrelu}
        \end{figure}

        \paragraph{Update histograms.}
        Figures \ref{fig:update_hist_relubn} and \ref{fig:update_hist_lrelu} present histograms of the absolute values of parameter updates for the optimizee with batch normalization and ReLU and for the optimizee with Leaky ReLU, respectively.

        We can again notice that the learned optimizer gradually changes the shape of its update distribution from one in which large values are predominant to a shape similar to that of Adam.
        
        %%% Update histograms

        % make the figure stay at the top of the last page
        % \null
        % \vfill

\end{document}